%% file: main.tex
\documentclass[letterpaper, 10pt, conference]{ieeeconf}  

\IEEEoverridecommandlockouts                              

\usepackage{graphicx}
\usepackage{amsmath}
\usepackage{colortbl}
\usepackage{amssymb}
\usepackage{booktabs}
\usepackage{lipsum} 
\usepackage[absolute,overlay]{textpos}
\usepackage{multirow}
\usepackage{url}
\usepackage{xcolor}
\usepackage{color}
\usepackage{pifont}
\usepackage{hyperref}
\usepackage{wrapfig}
\usepackage{multirow}
\usepackage{tabularx}
\usepackage{caption}

\usepackage{graphicx}
\usepackage{subcaption}

\usepackage{float}

\hypersetup{pdfstartview=FitH,
            colorlinks=true,
            linkcolor=red,
            anchorcolor=blue,
            citecolor=blue
            }

\usepackage{bbding}

\title{\LARGE \bf
TopoNav: Topological Graphs as a Key Enabler for Advanced Object Navigation
}

\author{Peiran Liu$^{1,2,*}$, Qiang Zhang$^{1,2,*}$, Daojie Peng$^{1,*}$, Lingfeng Zhang$^{3,*}$, \\ Yihao Qin$^{1}$, Hang Zhou$^{1}$, Jun Ma$^{1}$, Renjing Xu$^{1,\dagger}$, Yiding Ji$^{1,\dagger}$
\thanks{$^\dagger$Corresponding author; $^*$Equal contribution.}
\thanks{$^{1}$The authors are with The Hong Kong University of Science and Technology (Guangzhou), Guangzhou, China. {\tt\small pliu868@connect.hkust-gz.edu.cn}
}%
\thanks{{$^{2}$The authors are with Beijing Innovation Center of Humanoid Robotics Co., Ltd.}
{\tt\small  jony.zhang@x-humanoid.com}}
\thanks{{$^{3}$The authors are with Shenzhen International Graduate School, Tsinghua University.}}
}

\begin{document}

\maketitle

\input{tex/objnav_version/z0_abstrct}
\input{tex/objnav_version/z1_introduction}
\input{tex/objnav_version/z2_related_works}

\input{tex/z4_methdology}
\input{tex/z5_experiments}
\input{tex/z6_conclusion}


\bibliographystyle{IEEEtran}
\bibliography{ref}

\end{document}

%% file: tex/objnav_version/z0_abstrct.tex
\begin{abstract}

Object Navigation (ObjectNav) has made great progress with large language models (LLMs), but still faces challenges in memory management, especially in long-horizon tasks and dynamic scenes. To address this, we propose TopoNav, a new framework that leverages topological structures as spatial memory. By building and updating a topological graph that captures scene connections, adjacency, and semantic meaning, TopoNav helps agents accumulate spatial knowledge over time, retrieve key information, and reason effectively toward distant goals. Our experiments show that TopoNav achieves state-of-the-art performance on benchmark ObjectNav datasets, with higher success rates and more efficient paths. It particularly excels in diverse and complex environments, as it connects temporary visual inputs with lasting spatial understanding.

\end{abstract}

%% file: tex/objnav_version/z1_introduction.tex
\section{Introduction}
\label{sec:intro}

ObjectNav is a pivotal task in embodied AI and robotic interaction, requiring seamless integration of visual perception, natural language understanding, and spatial reasoning to enable agents to navigate in unknown environment to find specifed target object. In recent years, we have witnessed remarkable advancements in ObjectNav, largely driven by the rapid evolution of LLMs, which have significantly enhanced capabilities in instruction parsing, cross-modal alignment, and context-aware reasoning. State-of-the-art frameworks leveraging pre-trained LLMs and vision-language encoders have achieved impressive results in simple or constrained scenarios, underscoring ObjectNav’s potential as a cornerstone for intelligent agent-environment interaction.

\begin{figure}[ht]
  \centering
  \includegraphics[width=\linewidth]{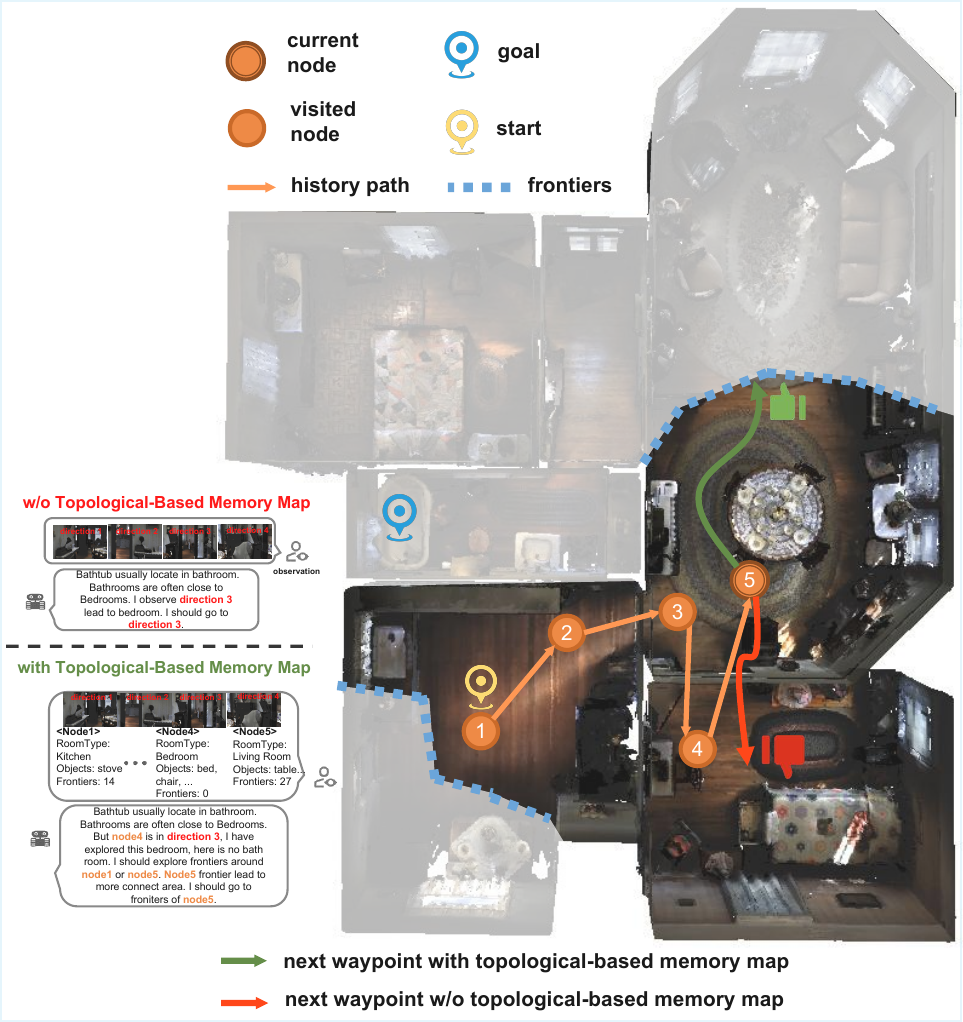}
  \caption{\small Our memorization technique enables backtracking to key nodes when exploration directions prove suboptimal, enhancing navigation efficiency through historical trajectory recall.}
  \label{fig:teaser}
  \vspace{-25pt}
\end{figure}

Despite these progresses, we observe that current ObjectNav methods still face critical bottlenecks, particularly in complex, dynamic, or large-scale environments. 

A key limitation lies in the lack of robust spatial memory mechanisms: most approaches rely heavily on ephemeral visual observations, struggling to accumulate and retain long-term structural knowledge of environments. This deficiency leads to fragmented memory of environmental structures, often resulting in goal confusion, redundant paths, or even complete navigation failure during long-horizon tasks—hindering the generalization of navigation systems to real-world, unconstrained scenarios.

Our key insight is that topological structures inherently function as a form of compact, enduring spatial memory. Unlike pixel-level visual details that are transient and data-heavy, topological information captures the essence of environmental structure in a stable, decision-supporting format. Such properties make topology uniquely suited to address navigation's memory deficits. While topological maps have long been used in traditional robotic navigation, we note their potential as dynamic memory carriers in ObjectNav—especially in conjunction with language-driven spatial reasoning—remains largely untapped, creating a critical research gap.

\begin{figure*}[t]
    \centering
    \includegraphics[width=0.98\textwidth]{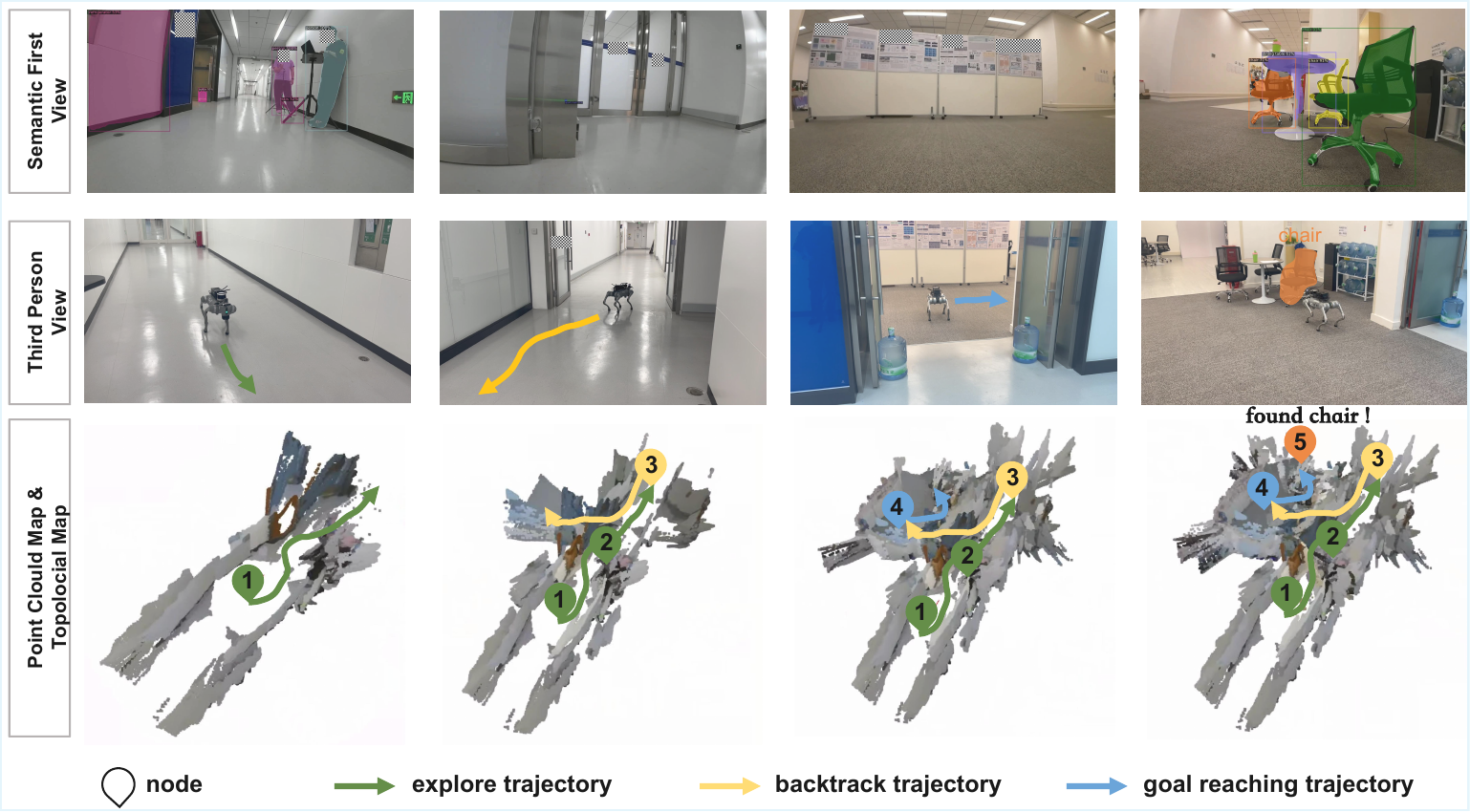}
    \caption{\small\textbf{Real Robot Experiment Demonstration}: When prolonged hallway exploration is detected, the robot backtracks to a node associated with a room and then starts a procedure inside the room to search for chairs.}
    \label{fig:go2_exp}
    \vspace{-20pt}
\end{figure*}

To address this gap, we introduce \textbf{TopoNav}, a novel framework that constructs and maintains a dynamic topological memory graph as the core of its navigation system. TopoNav’s key innovation is modeling environmental topology as actionable spatial memory: critical regions (nodes) and their connectivity (edges) are encoded in a graph structure that evolves in real time as the agent navigates. We design this topological memory graph to be tightly integrated with LLM-based instruction parsing and visual perception modules, enabling the agent to accumulate, update, and retrieve spatial knowledge dynamically—thus supporting coherent reasoning about long-range navigation goals and correcting deviations from optimal paths.

Our contributions are threefold:
\begin{enumerate}
\item We establish the theoretical connection between topological structures and spatial memory in ObjectNav, proposing a novel paradigm for enhancing navigation robustness.
\item We introduce \textbf{TopoNav}, a practical framework that implements dynamic topological memory graphs to bridge transient visual inputs and persistent spatial understanding.
\item We demonstrate that TopoNav achieves state-of-the-art performance on benchmark ObjectNav datasets, outperforming existing methods in success rate and path efficiency, with its performance further validated through extensive real-world experiments.
\end{enumerate}

%% file: tex/objnav_version/z2_related_works.tex
\section{Related Work}
\subsection{Object Navigation}

Existing approaches to object navigation can be broadly categorized into two paradigms: the first relies on learning-based methods, including reinforcement learning~\cite{ye2021efficient,deitke2022}, imitation learning~\cite{cai2024bridging}, and predictive models generating bird's-eye view maps~\cite{luostubborn, chaplot2020object, ramakrishnan2022poni, zhai2023peanut}, while the second employs zero-shot methods that leverage open-vocabulary scene understanding, eliminating the need for task-specific training—these zero-shot approaches are further divided into image-based methods projecting target objects into visual embedding spaces for matching~\cite{majumdar2022zson, gadre2023cows}, and map-based methods utilizing frontier-based exploration combined with LLMs for semantic reasoning and decision-making~\cite{yu2023l3mvn, yokoyama2024vlfm, zhang2024trihelper, zhang2024multi, zhang2025apexnav}. However, current approaches lacks explicit spatial memory mechanisms, leading to fragmented reasoning in complex scenarios. These limitations highlight the need for architectures that integrate LLMs with enduring spatial representations, such as topological memory graphs, to support coherent long-term planning.

\subsection{Memory Mechanisms in Navigation}

To address the memory limitations of LLM-based Vision Language Navigation (VLN), recent research has explored hierarchical memory systems and continual learning paradigms. The Dual Memory Networks propose a hybrid architecture combining static memory and dynamic memory \cite{zhang2024dual}. This approach mitigates catastrophic forgetting but remains constrained by predefined memory structures, failing to encode dynamic spatial relationships.

In the realm of continual learning, the CVLN paradigm introduces a sequential training framework for VLN agents, enabling them to adapt to new environments while retaining knowledge of previously learned scenes \cite{jeong2024continual}. However, CVLN focuses on incremental task adaptation rather than structural memory consolidation, making it ill-suited for environments requiring persistent topological understanding. Meanwhile, Mem4Nav presents a hierarchical spatial-cognition system that fuses sparse octrees with semantic topology graphs, achieving gains in task completion rates on large-scale urban datasets \cite{liu2025mem4nav}. This work underscores the importance of multiscale memory encoding but relies on predefined topological priors, limiting its applicability to unseen environments.
A critical gap persists in current memory mechanisms: most methods store data-level observations rather than structure-level relationships. TopoNav addresses this by modeling topology as actionable spatial memory, dynamically updating graph nodes and edges to reflect real-time environmental changes \cite{wei2024exploring}.

\subsection{Topological Structures for Spatial Navigation}

Topological maps have gained renewed attention in VLN, particularly for encoding environmental connectivity and long-term spatial relationships. The ETPNav framework introduces an online topological planning system that self-organizes waypoints into dynamic graphs \cite{an2024etpnav}. However, ETPNav’s topological maps are constructed based on precomputed waypoint graphs, lacking real-time adaptability to unseen environmental changes \cite{an2024etpnav}. Similarly, Revind combines offline reinforcement learning with topological graphs to enable long-horizon navigation in real-world scenarios \cite{shah2022offline}, but its reliance on predefined graph structures limits generalization to novel environments.

Recent studies have also explored GNN-based topological reasoning to enhance spatial understanding. For example, Su et al. analyze the topology awareness of GNNs and demonstrate their ability to preserve structural information in graph representations \cite{su2024topology}. However, these methods primarily focus on node classification tasks and have not been adapted for VLN’s multimodal reasoning requirements. In contrast, TopoNav integrates topological graphs with LLM-driven instruction parsing, enabling semantic-aware topological updates that dynamically align with linguistic cues \cite{wei2024exploring}.

Crucially, TopoNav distinguishes itself by treating topological structures as evolving memory carriers rather than static maps. Unlike prior work that uses topology for coarse path planning \cite{an2024etpnav, shah2022offline}, TopoNav’s dynamic topological memory graph continuously refines spatial knowledge through visual-linguistic interactions, supporting adaptive deviation correction and long-range goal reasoning in real time \cite{wei2024exploring}.

%% file: tex/z4_methdology.tex
\section{Methodology}
\label{sec:exp}

\subsection{Problem Formulation}
\label{sec:problem_formulation}
The ObjectNav task~\cite{batra2020objectnav} requires autonomous agents to locate and approach predefined target object categories in previously unexplored environments. Formally, given a set of target object categories $\mathcal{T} = \{T_1, T_2, \dots, T_N\}$ (e.g., ``chair'', ``bed'', etc,.), an episode initiates by instantiating an agent at a randomized starting pose within an unseen scene $\mathcal{S}$. The agent is assigned a specific target category $T_i \in \mathcal{T}$ as its navigation objective. At each discrete timestep $t$, the agent perceives an observation tuple $O_t = (V_t, P_t)$, where $V_t$ represents egocentric visual inputs comprising RGB and depth images, while $P_t$ denotes the agent's proprioceptive pose information (position and orientation). Based on the sensory inputs, the agent must navigate to a specified object $T_i$ in the unknown environment $\mathcal{S}.$ 

\subsection{Overview}
\label{sec:overview}
\begin{figure*}[ht]
  \centering
  \includegraphics[width=\textwidth]{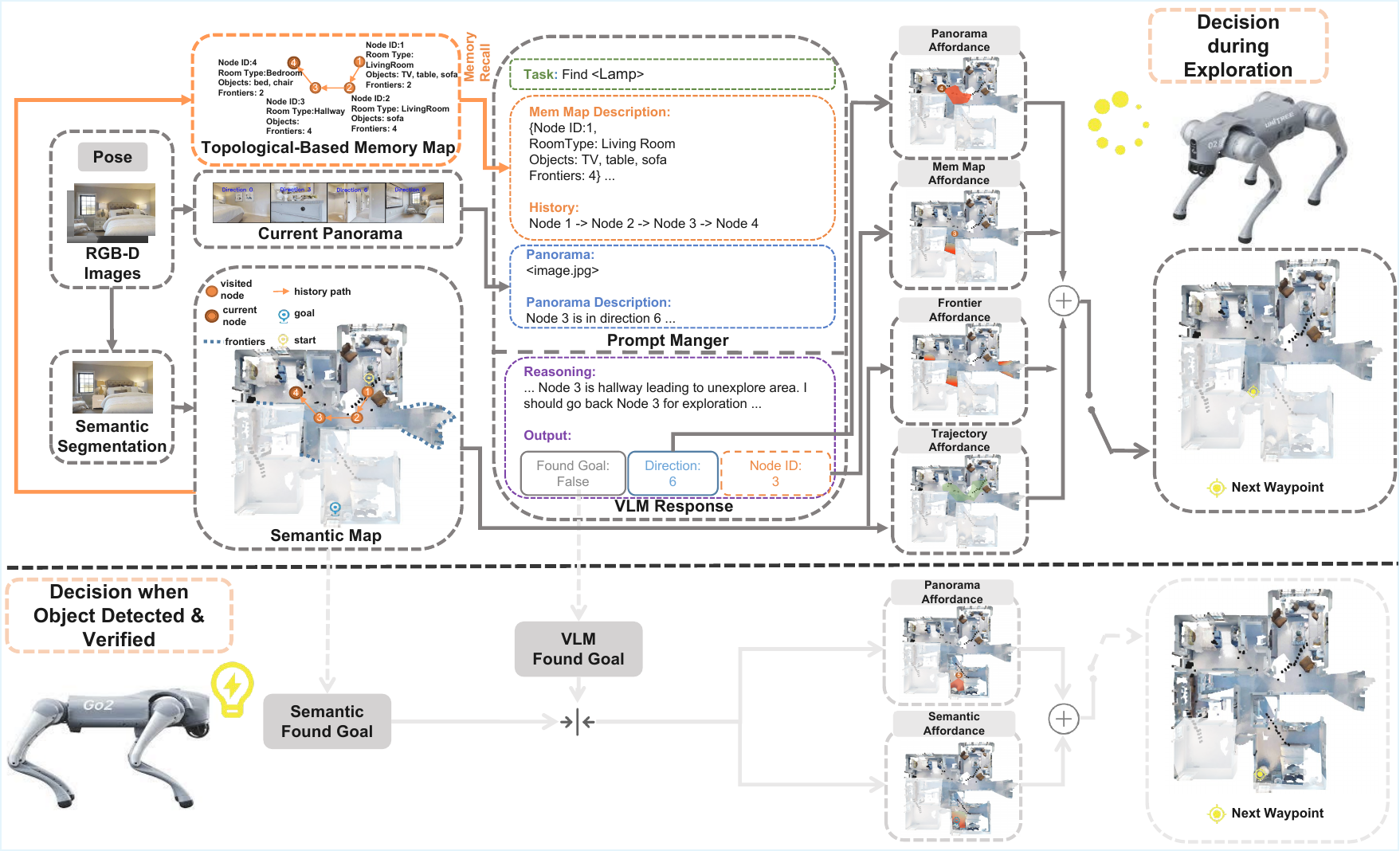}
  \caption{\small \textbf{Framework Overview}. Our approach constructs both a semantic point cloud map and a topological memory map during navigation. A prompt manager integrates the current observation with these map representations. Finally, our affordance-driven waypoint selection strategy dynamically adapts between exploration and target acquisition phases.}
  \label{fig:pipeline}
  \vspace{-18pt}
\end{figure*}

Our proposed framework architecture is illustrated in Fig.~\ref{fig:pipeline}. At each timestep, the system receives an RGB-D image $V_t$ and agent pose $P_t$ as inputs. We construct a semantic point cloud map following the methodology detailed in Section~\ref{sec:pcd_map}. Concurrently, we maintain a text-based topological memory map as described in Section~\ref{sec:topo_map}. These representations are integrated through our fused policy, which combines the semantic point cloud map, topological memory map, and Vision Language Model (VLM) guidance to select optimal waypoint candidates according to the strategy defined in Section~\ref{sec:waypoint_strategy}.

\subsection{Semantic Point Cloud Map Construction}
\label{sec:pcd_map}

The mapping module constructs a point cloud representation of the environment using visual observations $V_t$ and pose information $P_t$ defined in Section \ref{sec:problem_formulation}. This representation comprises five components: a comprehensive \textit{scene point cloud} $\mathcal{M}_{scene}$, semantically segregated \textit{object point clouds} $\mathcal{M}_{obj}$, a filtered \textit{navigable point cloud} $\mathcal{M}_{nav}$, a \textit{obstacle point cloud} $\mathcal{M}_{obs}$ and a \textit{frontier point cloud} $\mathcal{M}_{fro}$. The mapping process evolves incrementally throughout navigation, maintaining both geometric and semantic consistency across observations.

At each timestep $t$, the system processes RGB images $I_t \in \mathbb{R}^{H \times W \times 3}$ and depth maps $D_t \in \mathbb{R}^{H \times W}$ from the observation $V_t$. The agent's pose $P_t = (\mathbf{p}_t, \mathbf{q}_t)$ provides position $\mathbf{p}_t \in \mathbb{R}^3$ and orientation $\mathbf{q}_t$ (represented as quaternion). Using the known camera intrinsics matrix $\mathbf{K} \in \mathbb{R}^{3 \times 3}$, each valid depth pixel $(u,v)$ with depth value $d = D_t(u,v)$ is transformed to camera coordinates:
\vspace{-5pt}
\begin{equation}
\mathbf{x}_c = 
\begin{bmatrix} 
x_c \\ 
y_c \\ 
z_c 
\end{bmatrix} = 
d \cdot \mathbf{K}^{-1}
\begin{bmatrix} 
u \\ 
v \\ 
1 
\end{bmatrix}.
\vspace{-3pt}
\label{equ:pix2cam}
\end{equation}
The corresponding RGB value $\mathbf{c} = I_t(u,v)$ is associated with each generated 3D point. These camera-space points $\mathbf{x}_c$ are then transformed to the global coordinate frame via rigid body transformation:
\vspace{-5pt}
\begin{equation}
\mathbf{x}_w = \mathbf{R}(\mathbf{q}_t) \mathbf{x}_c + \mathbf{p}_t ,
\vspace{-3pt}
\label{equ:cam2world}
\end{equation}

where $\mathbf{R}(\mathbf{q}_t) \in \mathbb{R}^{3 \times 3}$ is the rotation matrix derived from the orientation quaternion $\mathbf{q}_t$.

The scene point cloud $\mathcal{M}_{scene}$ aggregates all observed points $\mathbf{x}_w$ with their associated color attributes $\mathbf{c}$. To reduce measurement noise, a voxel downsampling operation is applied after incorporating new points:
\vspace{-5pt}
\[
\mathcal{M}_{scene} \leftarrow \text{VoxelDownsample}\left( \mathcal{M}_{scene} \cup \{\mathbf{x}_w^{(i)} \}, r_{pcd} \right),
\vspace{-3pt}
\]
where $r_{pcd}$ specifies the voxel grid resolution. This operation replaces points within each cubic voxel of size $r_{pcd}$ with their centroid, ensuring uniform point density while preserving geometric features.

Semantic segmentation is applied to the RGB image $I_t$ to identify regions corresponding to interesting object categories in $\mathcal{T}$. The resulting object masks $M_{obj}$ select depth pixels belonging to recognized objects. For each object class $c \in \mathcal{T}$, the corresponding points in the depth map $D_t$ are projected and transformed using the same process outlined in Equations~\ref{equ:pix2cam} and~\ref{equ:cam2world}. These class-specific points are accumulated in separate object point clouds:
\vspace{-5pt}
\[
\mathcal{M}_{obj}^{(c)} \leftarrow \text{VoxelDownsample}\left( \mathcal{M}_{obj}^{(c)} \cup \{\mathbf{x}_w^{(c,j)}\}, r_{pcd} \right),
\vspace{-3pt}
\]
where $\mathbf{x}_w^{(c,j)}$ denotes the $j$-th point belonging to object class $c$. These class-specific representations enable efficient spatial querying of potential target objects $T_i$ during navigation.

The navigable point cloud $\mathcal{M}_{nav}$ is constructed through a sequential geometric and semantic refinement process to identify traversable regions. Initial candidate points are extracted based on height constraints relative to the estimated floor level, formalized as:
\vspace{-5pt}
\[
\mathcal{M}_{nav} = \{ \mathbf{x}_w \in \mathcal{M}_{scene} : z_{floor} - \delta \leq z_w \leq z_{floor} + \delta \},
\vspace{-3pt}
\]
where $z_{floor}$ represents the ground plane elevation, and $\delta$ accommodates floor irregularities. This geometrically filtered set is then augmented with semantically navigable structures: Points belonging to object classes in $\mathcal{C}_{nav} \subset \mathcal{T}$ (such as stairs) are incorporated regardless of height, yielding:
\vspace{-5pt}
\[
\mathcal{M}_{nav} \leftarrow \mathcal{M}_{nav} \cup \left( \bigcup_{c \in \mathcal{C}_{nav}} \mathcal{M}_{obj}^{(c)} \right).
\vspace{-3pt}
\]

To ensure spatial connectivity in sparse observation conditions, linear interpolation bridges gaps between the agent's standing position $\mathbf{p}_{stand} = [x_t, y_t, z_{floor}]$ and observed navigable points. For each $\mathbf{x}_i \in \mathcal{M}_{nav}$ with $\lVert \mathbf{x}_i - \mathbf{p}_{stand} \rVert_2 > \Delta_{step}$, intermediate points are generated:

\vspace{-12pt}
\begin{align*}
    &\mathbf{p}_{k} = \mathbf{p}_{stand} + \frac{k \cdot \Delta_{step}}{\lVert \mathbf{x}_i - \mathbf{p}_{stand} \rVert_2} (\mathbf{x}_i - \mathbf{p}_{stand}) \quad \\
    &\text{for} \quad k = 1, 2, \dots, \lfloor \lVert \mathbf{x}_i - \mathbf{p}_{stand} \rVert_2 / \Delta_{step} \rfloor    ,
\vspace{-3pt}
\end{align*}

retaining only those within the navigable height band $\left|z_k - z_{floor}\right| < \delta$. This interpolation establishes continuous traversable surfaces even with partial observations. We then downsample to uniform resolution:
\vspace{-5pt}
\[
\mathcal{M}_{nav} \leftarrow \text{VoxelDownSample}(\mathcal{M}_{nav}, r_{pcd}).
\vspace{-3pt}
\]

The obstacle point cloud $\mathcal{M}_{obs}$ is constructed by selecting all points located above the floor elevation. Formally: 
\vspace{-5pt}
\[
\mathcal{M}_{obs} = \left\{ \mathbf{x}_w \in \mathcal{M}_{scene} : z_w > z_{floor} + \delta \right\}.
\vspace{-3pt}
\]

The frontier point cloud $\mathcal{M}_{fro}$ identifies boundary regions between explored navigable areas and unknown space. This representation is constructed through geometric analysis of the navigable and obstacle point clouds within a discretized 2D grid with width $N_x$ and height $N_y$, where
\vspace{-5pt}
\[
N_x = \left\lceil \frac{x_{max} - x_{min}}{r_g} \right\rceil, \quad
N_y = \left\lceil \frac{y_{max} - y_{min}}{r_g} \right\rceil.
\vspace{-3pt}
\]
Each grid cell $\mathcal{G}(i,j)$ is populated by projecting points from $\mathcal{M}_{nav}$ and $\mathcal{M}_{obs}$:
\vspace{-5pt}
\[
\mathcal{G}(i,j) = \begin{cases} 
1 & \text{if } \exists \mathbf{p} \in \mathcal{M}_{nav}: \phi(\mathbf{p}) = (i,j) \\
-1 & \text{if } \exists \mathbf{p} \in \mathcal{M}_{obs}: \phi(\mathbf{p}) = (i,j) \\
0 & \text{otherwise},
\end{cases}
\vspace{-3pt}
\]
where $\phi(\mathbf{p}) = \left( \left\lfloor \frac{p_x - x_{min}}{r_g} \right\rfloor, \left\lfloor \frac{p_y - y_{min}}{r_g} \right\rfloor \right)$ with $r_g$ represent the grid resolution.

Then we select frontier candidates points $\mathcal{F}_{candidate}$ using ~\cite{yamauchi1997frontier}. Finally, we project 2D grid to 3D at floor height with
\vspace{-5pt}
\[
\mathbf{x}_{fro}^{(t)} = \begin{bmatrix} i_k \cdot r_g + x_{min} \\ j_k \cdot r_g + y_{min} \\ z_{floor} \end{bmatrix} \quad \forall (i_k, j_k) \in \mathcal{F}_{candidate}.
\vspace{-5pt}
\]
The resulting point cloud $\mathcal{M}_{fro}$ provides target locations for exploration while maintaining safe clearance from obstacles.

\subsection{Topological Memory Map Construction}
\label{sec:topo_map}

Complementing the geometric representation defined in Section \ref{sec:pcd_map}, we maintain a hierarchical topological map $\mathcal{M}_{topo} = \{\mathcal{N}_k\}_{k=1}^K$ as shown in Fig.~\ref{fig:3d_vis}. This abstraction transforms continuous environmental geometry into semantically meaningful nodes, enabling high-level reasoning essential for long-horizon navigation tasks. Each topological node $\mathcal{N}_k = \langle \text{id}_k, \mathbf{n}_k, \mathcal{O}_k, R_k, f_k \rangle$ contains five navigation-critical attributes:

\begin{figure}[ht]
  \centering
  \includegraphics[width=\linewidth]{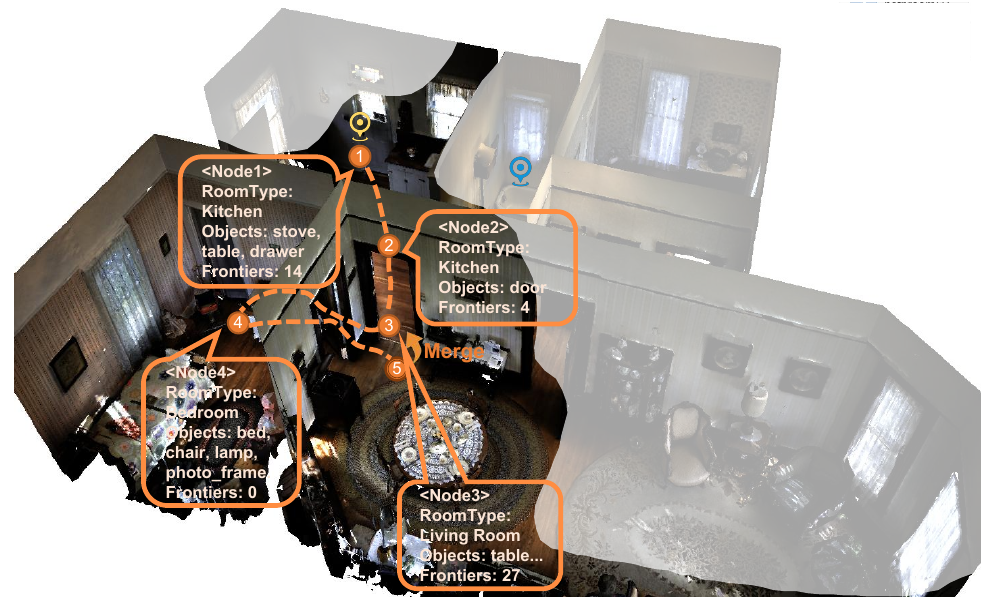}
  \caption{\small The text-based topological map enables VLMs to perform high-level reasoning and memory-based decisions.
 }
  \label{fig:3d_vis}
  \vspace{-15pt}
\end{figure}

\begin{enumerate}
    \item[1.] Node ID: a unique ID $\text{id}_k$ to distinguish nodes.  

    \item[2.] Position: $\mathbf{n}_k = (x_k, y_k)$ in world coordinates (z-coordinate omitted as topological reasoning operates primarily in 2D space).
    
    \item[3.] Surrounding object classes: $\mathcal{O}_k \subset \mathcal{T}$ identified within a $r_{topo}$ radius of $\mathbf{n}_k$. This semantic context informs object search strategies during target-driven navigation.
    
    \item[4.] Room type: $R_k$ derived from VLM analysis of panoramic RGB observations $V_t^{panorama}$. This high-level semantic classification enables room-aware navigation policies.
    
    \item[5.] Frontier count: $f_k = |\mathcal{F}_{\mathbf{n}_k}|$ where $\mathcal{F}_{\mathbf{n}_k} = \{ \mathbf{x}_w \in \mathcal{M}_{fro} : \|\mathbf{x}_w - \mathbf{n}_k\|_2 < r_{topo} \}$.
\end{enumerate}

The topological map initializes at episode start with $\mathcal{N}_1$ at the agent's starting position $\mathbf{p}_0$. New nodes are incrementally created when the agent arrive a new waypoint (determined by the strategy in Section \ref{sec:waypoint_strategy}). 

Node creation involves:
\vspace{-5pt}
\[
\mathcal{N}_{new} = \left\langle \text{id}_k, \mathbf{p}_k, \mathcal{O}_k, R_k, f_k \right\rangle
\vspace{-3pt}
\]
where:
\vspace{-5pt}
\begin{align*}
\mathcal{O}_k &= \left\{ c \in \mathcal{T} : \exists {\mathbf{x}_w \in \mathcal{M}_{obj}^{(c)}} \text{ s.t. } \|\mathbf{x}_w - \mathbf{n}_k\|_2 < r_{topo} \right\} \\
R_k &= \Psi_{\text{VLM}}(V_t^{panorama}) \\
f_k &= |\{ \mathbf{x}_w \in \mathcal{M}_{fro} : \|\mathbf{x}_w - \mathbf{n}_k\|_2 < r_{topo} \}|
\vspace{-3pt}
\end{align*}
with $\Psi_{\text{VLM}}$ classifying room types (e.g., "kitchen", "bedroom") from panoramic imagery.

Node merging occurs when spatial proximity and traversability conditions are satisfied, preventing redundant representation of connected spaces:
\vspace{-5pt}
\begin{gather*}
\|\mathbf{n}_i - \mathbf{n}_j\|_2 < d_{merge} \land  \\
\nexists \mathbf{x}_w \in \mathcal{M}_{obs} : \left( z_w \in [z_{floor}, z_{ceiling}] \right) \cap \left( \mathbf{x}_w \in \mathcal{L}(\mathbf{n}_i, \mathbf{n}_j) \right) ,
\vspace{-3pt}
\end{gather*}
where $\mathcal{L}(\mathbf{n}_i, \mathbf{n}_j)$ denotes the straight-line path between nodes, ($z_{ceiling}$, $z_{floor}$) defines the vertical obstacle band, and $d_{merge}$ represent merge distance. Merged nodes retain the earliest node ID and room type $R$ for temporal consistency, with updated attributes:
\[
\mathbf{n}' = \frac{\mathbf{n}_i + \mathbf{n}_j}{2}
\]
\[
\quad \mathcal{O}'_t = \left\{ c \in \mathcal{T} : \exists {\mathbf{x}_w \in \mathcal{M}_{obj}^{(c)}} \text{ s.t. } \|\mathbf{x}_w - \mathbf{n}'\|_2 < r_{topo} \right\}
\]
\[
f'_k = |\{ \mathbf{x}_w \in \mathcal{M}_{fro} : \|\mathbf{x}_w - \mathbf{n}\|_2 < r_{topo} \}|.
\]
The merged node position is the centroid of the two merging nodes, with surrounding objects $\mathcal{O}'$ and frontier count $f'$ recomputed from $\mathcal{M}_{obj}$ and $\mathcal{M}_{fro}$ respectively.

The topological map critically enhances navigation through three integrated capabilities: \textit{Semantic Goal Prioritization} leverages room-object correlations to guide target search (e.g., locating beds in bedrooms by aligning $T_i$ with $R_k$); \textit{Exploration Optimization} directs agents toward frontier-rich nodes to minimize redundant coverage; and \textit{History-Aware Planning} utilizes the navigation sequence $\mathcal{H} = (\mathcal{N}_1 \to \mathcal{N}_2 \to \cdots)$ to avoid recent revisits while permitting strategic returns to key nodes like hallways.

This hierarchical representation serves as a cognitive map bridging perception and action, enabling: efficient long-horizon planning, context-aware navigation through room-object associations, adaptive exploration based on frontier distributions, and memory-efficient environment modeling.

\subsection{Waypoint Selection Strategy}
\label{sec:waypoint_strategy}

The waypoint selection strategy integrates VLM guidance to determine optimal navigation targets. The VLM processes multimodal inputs including: 1) Target object category $T_{target}$, 2) Textual topological memory map representation $\mathcal{M}_{topo}$, 3) Navigation history $\mathcal{H}$, and 4) Current panoramic observation $V_t^{panorama}$. Based on these inputs, the VLM outputs three critical navigation parameters: 1) Next topological node ID $k'$, 2) Preferred direction $d_{vlm}$ within the panoramic view $V_t^{panorama}$, and 3) Binary target object detection indicator $g_{found} \in \{0,1\}$.

The VLM helps waypoint selection through an affordance-based framework operating over the navigable point cloud $\mathcal{M}_{nav}$. For each candidate point $\mathbf{p}_{nav}^i \in \mathcal{M}_{nav}$, we compute a composite affordance value $A(\mathbf{p}_{nav}^i)$ that integrates multiple navigation objectives. The strategy operates in two distinct phases based on target detection status.

The navigation affordances are computed using a unified framework. For any point $\mathbf{p}_i \in \mathcal{M}_{nav}$ and target point cloud $S$, we define:
\vspace{-5pt}
\[
d_i^S = \min_{\mathbf{q} \in S} \|\mathbf{p}_i - \mathbf{q}\|_2, \quad
d_{\min}^S = \min_i d_i^S, \quad
d_{\max}^S = \max_i d_i^S .
\vspace{-3pt}
\]
The normalized affordance is:
\vspace{-5pt}
\[
N(d_i^S) = 1 - \frac{d_i^S - d_{\min}^S}{d_{\max}^S - d_{\min}^S + \epsilon} \quad .
\vspace{-3pt}
\]
This normalization ensures affordance values range between 0 and 1, with higher values indicating more favorable positions.

\subsubsection{Exploration Phase}

When $g_{found} = 0$ (target not visually confirmed by VLM) and $T_{target} \notin \mathcal{O}$ (target object not in detected object list), the agent enters exploration mode. This phase employs four complementary affordance components to guide efficient exploration:

\textit{VLM Direction Affordance ($A_{dir}$)}: Incorporates the VLM's directional guidance by attracting the agent toward points aligned with the preferred direction:
\vspace{-8pt}
\[
A_{dir}(\mathbf{p}_{nav}^i) = N\left(d_i^{S_{dir}}\right)
,
\vspace{-5pt}
\]
where $S_{dir}$ is the point set along the VLM-selected direction $d_{vlm}$ in the panoramic view. This component leverages the VLM's visual reasoning capability to identify promising directions that potentially lead to the target object.

\textit{Node Affordance ($A_{node}$)}: Focuses exploration around frontiers near the VLM-selected topological node:
\vspace{-5pt}
\[
A_{node}(\mathbf{p}_{nav}^i) = N\left(d_i^{S_{node}}\right) ,
\vspace{-3pt}
\]
with $S_{node}$ representing frontier points associated with the VLM-selected node $\mathcal{N}_{k'}$. This affordance utilizes the VLM's topological reasoning ability to prioritize high-potential areas while avoiding repeated visits to the same regions.

\textit{Frontier Affordance ($A_{front}$)}: Encourages comprehensive exploration of unmapped regions:
\vspace{-5pt}
\[
A_{front}(\mathbf{p}_{nav}^i) = N\left(d_i^{\mathcal{M}_{fro}}\right) .
\vspace{-3pt}
\]
This component provides broad exploration pressure complementary to the more targeted VLM guidance.

\textit{History Avoidance Affordance ($A_{hist}$)}: Prevents redundant revisits to recently explored areas:
\vspace{-5pt}
\[
A_{hist}(\mathbf{p}_{nav}^i) = 1- N\left(d_i^{S_{hist}}\right) ,
\vspace{-3pt}
\]
where $S_{hist}$ contains trajectory history points. This mechanism promotes efficient exploration by encouraging novelty-seeking behavior.

\subsubsection{Target Acquisition Phase}

When $g_{found} = 1$ and target objects are confirmed by the detection model ($T_{target} \in \mathcal{O}$), the strategy transitions to target approach mode. This phase employs two key affordances:

\textit{VLM Direction Affordance ($A_{dir}$)}: Maintains continuity with exploration by retaining the VLM's directional guidance, ensuring consistent navigation toward the target area.
    
\textit{Semantic Proximity Affordance ($A_{sem}$)}: Directs the agent toward visually confirmed target instances:
\vspace{-5pt}
\[
A_{sem}(\mathbf{p}_{nav}^i) = N\left(d_i^{S_{sem}}\right),
\vspace{-3pt}
\]
where $S_{sem}$ is the set of points classified as the target object $T_{target}$ in $\mathcal{M}_{obj}$. This affordance enables precise approach to identified targets.

\subsubsection{Safety Integration and Waypoint Selection}

The composite affordance combines phase-specific components through summation:
\vspace{-5pt}
\[
A(\mathbf{p}_{nav}^i) = 
\begin{cases} 
\sum A_{\{dir,node,front,hist\}} & \text{exploration} \\
A_{dir} + A_{sem} & \text{target acquisition},
\end{cases}
\vspace{-3pt}
\]

Obstacle avoidance is implemented by zeroing out affordances near obstacles:
\vspace{-5pt}
\[
A_{obs}(\mathbf{p}_{nav}^i) = N\left(d_i^{\mathcal{M}_{obs}}\right)
\vspace{-3pt}
\]
\vspace{-12pt}
\[
A(\mathbf{p}_{nav}^i) = A(\mathbf{p}_{nav}^i) \cdot \mathbb{I}(A_{obs}(\mathbf{p}_{nav}^i) > \sigma),
\vspace{-3pt}
\]
where $\sigma$ is a safety threshold parameter determining the minimum safe distance from obstacles. This operation eliminates navigation options that are too close to potential hazards.

The target waypoint is selected through global maximization over the navigable space:
\vspace{-5pt}
\[
\mathbf{p}_{target} = \underset{\mathbf{p}^i \in \mathcal{M}_{nav}}{\arg\max} \, A(\mathbf{p}_{nav}^i).
\vspace{-3pt}
\]
This optimization ensures the agent moves toward positions that best balance exploration efficiency, target approach, and safety considerations.

\begin{table}[ht]
\scriptsize
    \centering
    \setlength{\tabcolsep}{1.4mm}{
    \begin{tabular}{cccccccc}
    \toprule[1.0pt]
    
        \multirow{2}{*}{\textbf{Method}} & \multirow{2}{*}{\textbf{Training Free}} & \multicolumn{3}{c}{\textbf{HM3D}}  & \multicolumn{3}{c}{\textbf{MP3D}} \\ \cmidrule(r){3-5} \cmidrule(l){6-8} 
        ~ & ~ & SR$\uparrow$ & SPL$\uparrow$ &DTG$\downarrow$ & SR$\uparrow$ & SPL$\uparrow$ & DTG$\downarrow$ \\ \midrule[1.0pt]
        SemExp\cite{chaplot2020object} & \XSolidBrush & - & - & - & 0.144 & 0.360 & 6.73 \\
        ZSON\cite{majumdar2022zson} & \XSolidBrush & 0.255 & 0.126 & - & 0.153 & 0.048 & - \\ 
        PONI\cite{ramakrishnan2022poni} & \XSolidBrush & - & - & - & 0.318 & 0.121 & 5.1 \\
        PixNav\cite{cai2024bridging} & \XSolidBrush & 0.379 & 0.205 & - & - & - & - \\ 
        SPNet\cite{zhao2023semantic} & \XSolidBrush & 0.312 & 0.101 & - & 0.163 & 0.048 & - \\ 
        SGM\cite{zhang2024imagine} & \XSolidBrush & \textbf{0.602} & 0.308 & - & 0.377 & 0.147 & 4.93 \\
        \midrule[1.0pt]
        
        CoW\cite{gadre2023cows} & \Checkmark  & - & - & - & 0.074 & 0.037 & -\\ 
        ESC\cite{zhou2023esc} & \Checkmark  & 0.392 & 0.223 & - & 0.287 & 0.142 & - \\

        VLFM\cite{yokoyama2024vlfm} & \Checkmark & 0.525 & 0.304 & - & 0.364 & \textbf{0.175} & - \\ 
        
        VoroNav\cite{wu2024voronav} & \Checkmark  & 0.420 & 0.260 & - & - & - & - \\
        L3MVN\cite{yu2023l3mvn}& \Checkmark  & 0.504 & 0.231 & 4.43 & - & - & - \\
        TriHelper\cite{zhang2024trihelper}& \Checkmark  & 0.565 & 0.253 & 3.87 & - & - & - \\ 
        InstructNav\cite{long2025instructnav}& \Checkmark  & 0.510 & 0.187 & \underline{2.89} & 0.420 & 0.161 & \underline{4.26} \\ 
        GAMap\cite{huang2024gamap}& \Checkmark  & 0.531 & 0.260 & - & - & - & - \\
        UniGoal\cite{yin2025unigoal}& \Checkmark  & 0.545 & 0.251 & - & 0.410 & 0.164 & - \\
        WMNav\cite{nie2025wmnav}& \Checkmark  & 0.581 & \underline{0.312} & - & \underline{0.454} & \underline{0.172} & - \\
        \midrule[1.0pt]
        \rowcolor{blue!8}\textbf{Ours} & \Checkmark & \underline{0.601} & \textbf{0.346} & \textbf{2.49} & \textbf{0.455} & 0.168 & \textbf{4.21} \\ 

    \bottomrule[1.0pt]
    \end{tabular}}
    \caption{\small Comparison with methods on HM3D and MP3D object goal navigation. We use \textbf{bold} and \underline{underline} to denote the first and second best performance respectively.   }\label{tab:comparison_experiments}
    \vspace{-10pt}
\end{table}

%% file: tex/z5_experiments.tex
\section{Experiments}

\subsection{Experiment Setup}
We evaluate our method in the Habitat Simulator~\cite{szot2021habitat} using HM3D and MP3D datasets. HM3D~\cite{ramakrishnan2021hm3d} comprises 2000 episodes across 20 scenes with 6 target objects. MP3D~\cite{Matterport3D} contains 2195 episodes spanning 11 scenes and 21 target objects. Our point cloud map construction builds upon the codebase of InstructNav~\cite{long2025instructnav}. We employ GPT-4o~\cite{hurst2024gpt} as the VLM and GLEE~\cite{wu2024general} as the vision foundation model for semantic segmentation. Although the simulation environment has limited target objects, GLEE's open-vocabulary capability enables generalization to open-vocabulary object navigation tasks. Experiments were conducted on a server with an RTX3090 GPU to support efficient detection model inference and parallel point cloud processing.

\subsection{Metrics}
We adopt three established metrics~\cite{anderson2018evaluation}: Success Rate (SR), Success weighted by Path Length (SPL), and Distance to Goal (DTG). Higher SR and SPL values indicate better performance. SPL combines task completion and path efficiency. DTG measures the final distance between agent and target at episode termination, where lower values are desirable.

\subsection{Evaluation}
\begin{figure}[h]
  \centering
  \includegraphics[width=\linewidth]{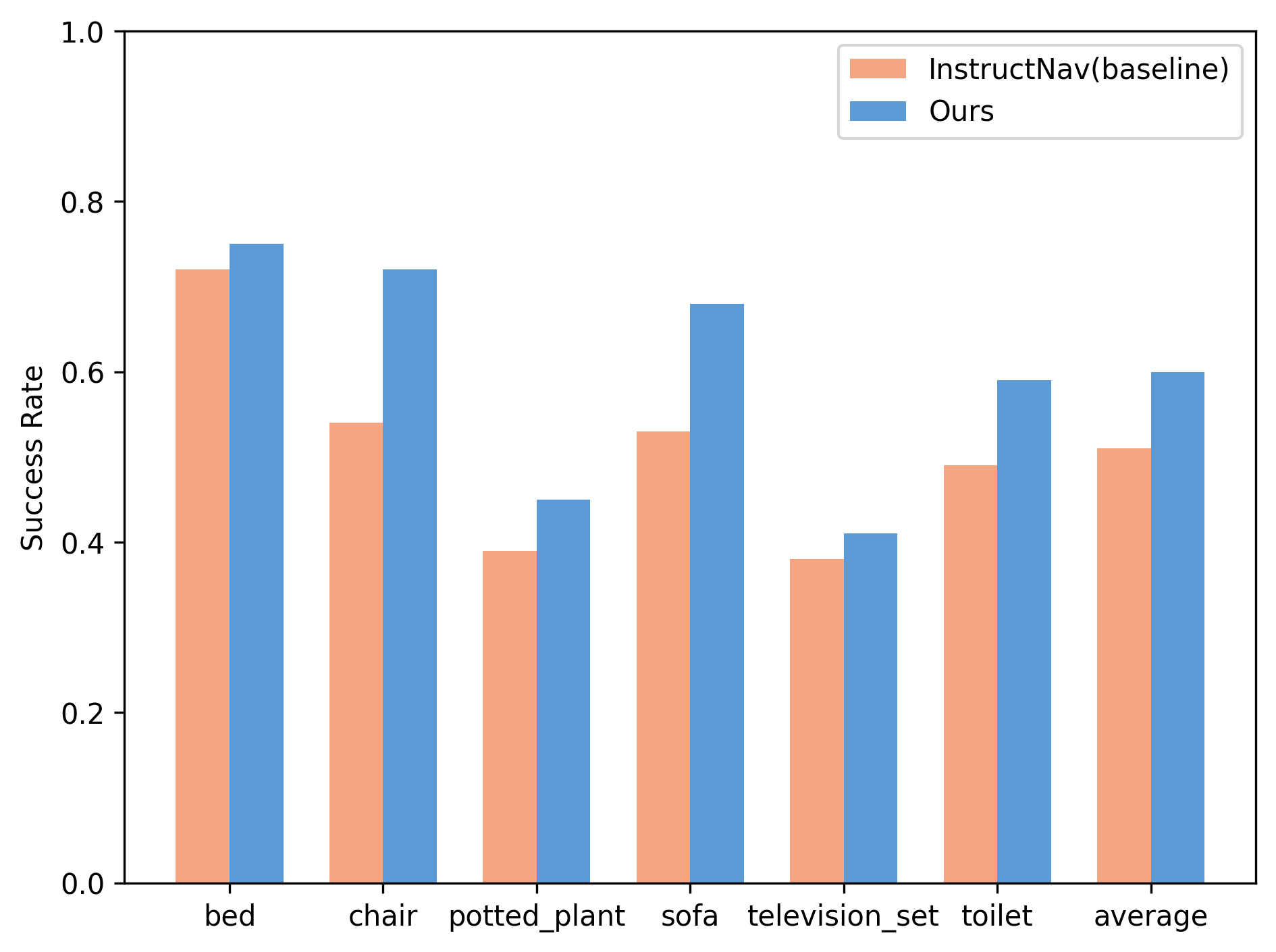}
  \caption{\small Camparison with baseline (InstructNav\cite{long2025instructnav}) on various objects.}
  \label{fig:category_comparison}
  \vspace{-15pt}
\end{figure}

Comparative experiment results in Habitat are presented in Table \ref{tab:comparison_experiments}. Our method achieves state-of-the-art (SOTA) performance for training-free ObjectNav on both datasets. On HM3D, our 0.601 SR exceeds the best training-free baseline WMNav by +2.0\%, while SPL improves by +10.9\% over VLFM. On MP3D, our SR surpasses WMNav by +0.2\%, and our 4.21 DTG sets a new benchmark. Remarkably, this training-free approach matches or exceeds leading training-based methods. Our HM3D 0.601 SR approaches SGM's training-based SOTA (0.602), while on MP3D we substantially outperform SGM's 0.377 SR by +20.7\%. These results demonstrate our method bridges the performance gap between training paradigms. As shown in Fig. \ref{fig:category_comparison}, TopoNav consistently outperforms InstructNav across all navigation tasks including bed, chair, potted plant, etc.

\subsection{Ablation Study}

\textbf{VLM and GLEE.} Both VLM and GLEE are quite important for better performance. The combination of VLM + GLEE demonstrates a higher SR and SPL compared to the individual methods only with VLM and only with GLEE (Table. \ref{tab:goal_detection_ablation}). This suggests that the integration of these two components enhances the overall success and efficiency of goal detection in terms of these metrics. Moreover, a smaller DTG of the VLM + GLEE method indicates that the combination also improves the spatial performance, enabling the robot to stop closer to the goal. 

Overall, the results consistently show that integrating VLM and GLEE yields a more robust and efficient goal detection framework for object navigation tasks, outperforming the use of either component in isolation across all evaluated metrics. These results highlight the complex interactions between different components in goal detection for object navigation.


\begin{table}[ht]
    \centering
    \setlength{\tabcolsep}{0.5em} 
    \renewcommand{\arraystretch}{1.2} 
    \begin{tabular*}{\linewidth}{@{\extracolsep{\fill}}lccc}
        \toprule
        \multirow{2}{*}{Method} & \multicolumn{3}{c}{Object Nav. on HM3D} \\
        \cmidrule{2-4}
         & \multicolumn{1}{c}{SR$\uparrow$} & \multicolumn{1}{c}{SPL$\uparrow$} & \multicolumn{1}{c}{DTG$\downarrow$} \\
        \midrule
        VLM + GLEE & 0.601 & 0.346 & 2.49 \\
        VLM only & 0.472 & 0.233 & 3.58 \\
        GLEE only & 0.445 & 0.223 & 3.39 \\
        \bottomrule
    \end{tabular*}
    \caption{\small Goal detection ablation.}
    \label{tab:goal_detection_ablation}
\vspace{-8pt}
\end{table}

\textbf{Node Information.} We also highlight the importance of topological node information. As shown in Table \ref{tab:node_ablation}, comparing with the topological node with all information, omitting any of these topological map node information types (Frontiers, Room Type, Object Type) causes a decline in SR and SPL, and an increase in DTG. 
This ablation study underscores the importance of topological map node information for robust object navigation.

\begin{table}[ht]
    \centering
    \setlength{\tabcolsep}{0.5em} 
    \renewcommand{\arraystretch}{1.2} 
    \begin{tabular*}{\linewidth}{@{\extracolsep{\fill}}lccc}
        \toprule
        \multirow{2}{*}{Method} & \multicolumn{3}{c}{Object Nav. on HM3D} \\
        \cmidrule{2-4}
         & \multicolumn{1}{c}{SR$\uparrow$} & \multicolumn{1}{c}{SPL$\uparrow$} & \multicolumn{1}{c}{DTG$\downarrow$} \\
        \midrule
        Current Memory & 0.601 & 0.346 & 2.49 \\
        w/o Frontiers & 0.544 & 0.267 & 2.70 \\
        w/o Room Type & 0.510 & 0.236 & 3.12 \\
        w/o Objects Type & 0.503 & 0.243 & 2.94 \\
        \bottomrule
    \end{tabular*}
    \caption{\small Topological map node information ablation.}
    \label{tab:node_ablation}
\vspace{-12pt}
\end{table}

\subsection{Real Robot Implementation}
Our experiments are conducted on a Unitree Go2 quadruped robot, where the full TopoNav system runs on a Jetson Orin. We use an Intel Realsense D435i for point cloud construction and a HESAI XT-6 lidar for SLAM. For semantic segmentation, we use GLEE~\cite{wu2024general}, a SOTA detection and segmentation model, and employ GPT-4o~\cite{hurst2024gpt} to maintain topological nodes and determine navigation directions. To build the semantic point cloud, we capture 9 RGBD panorama observations in real-world experiments (12 in simulation). As shown in Fig.~\ref{fig:go2_exp}, Go2 successfully explores a hallway, backtracks to the correct node, and navigates into a room to find the target object.

%% file: tex/z6_conclusion.tex
\section{Conclusion}
In this work, we present TopoNav, a zero-shot ObjectNav framework powered by topological mapping. By integrating a semantic point cloud with a graph-structured memory map, the system dynamically builds and updates spatial representations in real time, without prior training or scene-specific tuning. Nodes represent semantic regions and edges encode navigational connectivity, enabling efficient knowledge accumulation, path correction, and robust decision-making in novel environments. TopoNav achieves SOTA ObjectNav performance and demonstrates strong practicality in real-world experiments.